\documentclass[10pt,twocolumn,letterpaper]{article}

\usepackage[pagenumbers]{cvpr} % To force page numbers, e.g. for an arXiv version

\usepackage{amssymb}
\usepackage{comment}
\usepackage{amsfonts}       % blackboard math symbols
\usepackage{nicefrac}       % compact 
\usepackage{booktabs} % For professional looking tables (\toprule, \midrule, \bottomrule)
\usepackage{siunitx}  % For aligning numbers by decimal points and handling units
\usepackage{comment}
\usepackage{xstring} 
\usepackage{etoolbox}
\usepackage{algorithm}
\usepackage{tabularx}

\usepackage{graphicx}
\usepackage{amsmath}   % For writing great equations
\usepackage{amssymb}   % For great symbols
\usepackage{mathtools} % Extends amsmath
\usepackage{amsthm}    % For theorems, definitions, etc.
\usepackage{newtxtext}
\usepackage[cmintegrals]{newtxmath}
\usepackage{bm}

\definecolor{cvprblue}{rgb}{0.21,0.49,0.74}
\usepackage[pagebackref,breaklinks,colorlinks,allcolors=cvprblue]{hyperref}

\def\paperID{*****} % *** Enter the Paper ID here
\def\confName{CVPR}
\def\confYear{2026}

\title{Object-Aware 4D Human Motion Generation}

\author{
    Shurui Gui$^{\ast \, 1, \,2}$\thanks{Work done during internship at NEC Laboratories America.} \quad
    Deep Patel$^{\ast \, 2}$ \quad
    Xiner Li$^{1}$ \quad
    Martin Renqiang Min$^{2}$
    \\[2mm]
    $^{1}$Texas A\&M University \qquad
    $^{2}$NEC Laboratories America
    \\
    {\tt\small \{shurui.gui, lxe\}@tamu.edu \quad \{dpatel, renqiang\}@nec-labs.com}
    \\ 
    \centerline{\footnotesize $^{\ast}$Equal contribution.}
}

\begin{document}
\maketitle

\begin{abstract}
  Recent advances in video diffusion models have enabled the generation of high-quality videos. However, these videos still suffer from unrealistic deformations, semantic violations, and physical inconsistencies that are largely rooted in the absence of 3D physical priors. To address these challenges, we propose an object-aware 4D human motion generation framework grounded in 3D Gaussian representations and motion diffusion priors. With pre-generated 3D humans and objects, our method, Motion Score Distilled Interaction (MSDI), employs the spatial and prompt semantic information in large language models (LLMs) and motion priors through the proposed Motion Diffusion Score Distillation Sampling (MSDS). The combination of MSDS and LLMs enables our spatial-aware motion optimization, which distills score gradients from pre-trained motion diffusion models, to refine human motion while respecting object and semantic constraints. Unlike prior methods requiring joint training on limited interaction datasets, our zero-shot approach avoids retraining and generalizes to out-of-distribution object aware human motions. Experiments demonstrate that our framework produces natural and physically plausible human motions that respect 3D spatial context, offering a scalable solution for realistic 4D generation.
\end{abstract}

\section{Introduction}

Recent advancements in video generation have led to impressive results in generating realistic and semantically rich visual content. Video diffusion models~\citep{luo2023videofusion, bar2024lumiere, wu2023tune, chen2023videocrafter1, chen2024gentron, ma2024latte, zhou2022magicvideo} have achieved high visual quality on diverse tasks. Despite the progress, state-of-the-art models, including large-scale systems like Sora~\citep{liu2024sora}, still face persistent challenges such as unrealistic deformation, object penetration, and semantic violations. These issues often stem from the lack of explicit physical and spatial constraints, which are difficult to capture in purely 2D representations~\citep{Bahmani2022, Xie2023}.

To address these limitations, there has been a growing interest in incorporating 3D priors into generative modeling. The success of methods like DreamFusion~\citep{Poole2022} has demonstrated that distilling 2D priors from pre-trained diffusion models can guide 3D content generation, which has motivated many 3D and 4D generation works~\citep{Liu2023, Liu2023a, Deitke1314, Zhu2023, Zeng2024, Chu2024, Pan2024, Lee2024}. However, 4D generation methods that rely solely on video diffusion models inherit the spatial ambiguity and semantic misalignment problems. For instance, prompts involving spatial relations (e.g., “a dog under the bed”) often produce incorrect visual arrangements. To mitigate this, compositional 4D generation approaches~\citep{bahmani20244d, Xu2024, bahmani2024tc4d} have been proposed to combine multiple priors and synthesize novel distributions. Yet, these methods still face a fundamental bottleneck: human motions generated from pre-trained video models often suffer from distortions and fail to respect the physical constraints of interactions with static objects.

In this work, we tackle the challenging problem of zero-shot object-aware 4D human motion generation. Specifically, we aim to generate realistic 3D human motion interacting with a 3D static object over time, without requiring additional training on paired human-object data. Unlike prior methods~\citep{Cen2024, Wang2022, Xu2023, Diller2024} that rely on training dedicated models with limited joint human-object datasets, our framework leverages a compositional approach with strong generalization capability. Our method, Motion Score Distilled Interaction (MSDI), builds on recent advances in 3D Gaussian representations, i.e., motion diffusion models, and spatial reasoning with large language models (LLMs). 

Specifically, we first generate high-fidelity human and object 3D Gaussians using HumanGaussian~\citep{Liu2024} and DreamGaussian~\citep{Tang2023}, respectively. To control the temporal motion, we propose to guide human trajectories using LLM-generated spatial instructions, which provide coarse but plausible global motion plans. Then, instead of directly sampling from pre-trained motion diffusion models, which are often unreliably out of distribution, MSDS distills guidance from the motion diffusion model to form an optimization process that adjusts human poses and trajectories to align with both learned motion priors and interaction constraints. Furthermore, we formulate a constrained optimization framework that combines MSDS loss with smoothness, trajectory alignment, and collision-avoidance terms. This allows us to generate motion sequences that are realistic, smooth, and physically plausible with respect to the static object. Our zero-shot formulation ensures that the system can benefit from future improvements in motion diffusion models without the need for retraining, offering a scalable path toward generalizable and realistic object-aware 4D human motion generation. Experiments on multiple zero-shot prompts show, that our generated 4D scenes produce realistic motions with high physical constraint obedience ability while previous 4D generation methods can only generate unnatural distortions without plausible motion.

\section{Related Work}

\textbf{Video generation.} Video generation models have been widely used to generate realistic videos. Although the video diffusion models~\citep{luo2023videofusion, bar2024lumiere, wu2023tune, chen2023videocrafter1, chen2024gentron, ma2024latte, zhou2022magicvideo} have shown promising results in various areas, unrealistic deformation, twisted, penetration, and semantics violations still exist even in large video generation model Sora~\citep{liu2024sora}. These issues are often considered as the lack of physics information learned~\citep{yang2025towards, wang2022learning, bahmani20223d}. Despite many studies in addressing these issues by using trajectory tracking~\citep{geng2024motion}, occlusion masks~\citep{ke2021occlusion}, and semantic masks~\citep{pan2019video}, we argue that it is not feasible to solve this problem in 2D space without introducing extra information equivalent to physical information in 3D space, and the natural and intrinsic way to tackle these challenges should lie in the use of 3D space.

\textbf{3D generation.} While 3D generation has been explored in recent years~\citep{Tang2023, Shi2023, Liu2023, Liu2023a, Deitke1314, Zhu2023, Voleti2024}, one of the most popular and convincing directions is to extract prior knowledge from 2D diffusion models. Specifically, DreamFusion ~\citep{Poole2022}, the first method introducing Score Distillation Sampling (SDS), generates 3D content by leveraging information from 2D image diffusion models. This work inspires a significant amount of following works~\citep{Chen2024, Wu2024, Yi2024} on improving 3D content quality, optimization efficiency, and human avatar generations~\citep{Hu2024, Liu2024, Kocabas2024}. 

\textbf{4D Generation.} Aligning with the philosophy of extracting information from pre-trained image diffusion models, many 4D generation works adopt pre-trained video diffusion models~\citep{Zhang2023b, Liang2024, Yang2024}, to tackle challenges in image to 4D~\citep{Ren2023} and video to 4D~\citep{Yin2023, Luiten2023, Jiang2023, Zeng2024, Chu2024, Pan2024, Lee2024, Lei2024} tasks. However, the generation ability of these studies cannot go beyond the original distribution of the pre-trained diffusion models, and shares the same limitations as the original 2D pre-trained diffusion models. For example, most of the pre-trained video diffusion models face difficulties in understanding spatial information, \eg, generating with a prompt "a human walks towards the table" can produce unrealistic results, such as deformed bodies, poor framing that shows only the legs, or the human being omitted from the scene entirely. In order to solve this challenge, one convincing direction is to apply compositional 4D generation, which incorporates multiple prior distributions and combine them to generate samples with novel distributions. Recently, 4DFy~\citep{bahmani20244d}, Comp4D~\citep{Xu2024}, and TC4D~\citep{bahmani2024tc4d} have shown promising results on 4D compositional generation. However, although generating contents in 3D space helps with spatial information/trajectory planning, all the motion information from the pre-trained video diffusion model inherits its original distortions, especially on human-related motions. Motivated by this problem, we consider distilling information from dedicated motion models to guide the motion optimization process.

Different from interaction generations between humans and objects/scenes~\citep{Cen2024, Wang2022, Xu2023}, our method is zero-shot, which does not need to train specific dedicated models; thus, it widens the application range. Human-object interaction generations like InterDiff~\citep{Xu2023} and CG-HOI~\citep{Diller2024} require the joint distribution of humans and objects for training, while the sizes of these datasets are still limited, which cannot extend to any out-of-distribution scenarios. Since our work focuses on the interaction between humans and static objects, our setting is more similar to the HUMANISE task~\citep{Wang2022, Cen2024}. Compared with them, our setting eliminates the object-locating phase and focuses on the human trajectory and human motion generation with the static object. While these works require training extra models for human trajectory and human motion generation just for the additional object, our method does not require any additional motion diffusion model training and can achieve realistic interactions between humans and static objects with motion diffusion model score distillation sampling (MSDS). This zero-shot behavior enables this framework to improve as the motion diffusion model iterates in the future without extra distribution and retraining requirements.

\section{Preliminaries}

\subsection{3D Gaussian Splatting}

3D Gaussian Splatting~\citep{Kerbl2023} (3DGS) is a dominating representation in the 3D field, due to its explicit 3D space representation and high efficient optimization. The individual units of 3DGS are 3D Gaussian ellipsoids, where each 3D Gaussian is parameterized by position $\mu$, anisotropic covariance $\Sigma$ as its shape, and opacity $\alpha$ and spherical harmonic coefficients $sh$ as its optical characteristics, where $sh$ is a view-dependent property. The shape of the 3D Gaussian $\Sigma$ can be considered as the composition of a scaling and a rotation as follows:
\begin{equation}
    \Sigma = RSS^TR^T,
\end{equation}
where the scaling matrix $S$ can be denoted as a 3D vector $s$, and the rotation matrix $R$ as a quaternion $q\in \mathbf{SO}(3)$.

Therefore, the formal definition of a Gaussian centered at point $\mu$ is:
\begin{equation}
    G(\mathbf{x}, \mu)=e^{-\frac{1}{2}(\mathbf{x}-\mu)^T \Sigma^{-1}(\mathbf{x}-\mu)},
\end{equation}
where $\mathbf{x}$ is a random variable in 3D space.

To render 3D Gaussians into a 2D image, 3DGS considers the additional opacity $\alpha$ and spherical harmonic coefficients by utilizing a tile-based rasterizer and point-based $\alpha$-blend rendering. For each pixel $u$, its color $C(u)$ is rendered under the following calculation:
\begin{equation}
\begin{aligned}
    C(u)=\sum_{i \in N} T_i c_i \alpha_i \mathcal{S H}\left(s h_i, v\right),\\
    \quad T_i=G\left(\mathbf{x},
    \mu_i\right) \prod_{j=1}^{i-1}\left(1-\alpha_j G\left(\mathbf{x}, \mu_j\right)\right),
\end{aligned}
\end{equation}

where $T_i$ denotes the transmittance for the $i$-th Gaussian, $\mathcal{SH}$ denotes the spherical harmonic function, and $v$ represents the viewing direction. The 3D Gaussian optimization process includes adjusting all 3D Gaussian properties $\{\mu,q, s, \sigma, c\}$ and the high-level 3D Gaussian density modifications using densifying and pruning processes.

\subsection{SMPL-X} 

SMLP models~\citep{SMPLX} represent a human by transforming the human mesh of a standard pose, the canonical model, into the observation space, using pose parameter $\theta$, shape parameter $\beta$, and expression parameter $\phi$:
\begin{equation}
    \begin{aligned}
M(\beta, \theta, \phi) & =\operatorname{LBS}(T(\beta, \theta, \phi), J(\beta), \theta, \mathcal{W}), \\
T(\beta, \theta, \phi) & =\mathbf{T}+B_s(\beta)+B_e(\phi)+B_p(\theta),
\end{aligned}
\end{equation}
where $M$ is the function mapping parameters to a transformed human mesh model; $T$ represents the transformed human key points/vertices adjusted by different human shapes, expressions, and poses through corresponding functions $B_s$, $B_e$, and  $B_p$, respectively. Given the transformed vertices, the skins of the human mesh need to be adjusted according to the transformations of several nearby joints, which is done by the linear blend skinning function $\operatorname{LBS}(\cdot)$ where $\mathcal{W}$ stands for blend weights that determine the effects from different joints. Specifically, the LBS process is defined as follows:
\begin{equation}
    \mathbf{v}_o=\mathcal{G} \cdot \mathbf{v}_c, \quad \mathcal{G}=\sum_{k=1}^K w_k \mathcal{G}_k\left(\theta, j_k\right),
\end{equation}
where the vertices $\mathbf{v}_o$ in the observation space is deformed from the canonical pose vertices $\mathbf{v}_c$ by the deformation $\mathcal{G}$. The deformation is determined by the affine deformation $\mathcal{G}_k\left(\theta, j_k\right)$ that merges the warping effects from K neighboring joints, simulating the smooth position changes of vertices.

\section{Motion Score Distilled Interaction}\label{sec:method}

Our object-aware human motion generation (OAHM) framework addresses the challenging zero-shot generation problem by leveraging pre-trained motion priors within an explicit optimization paradigm. In this section, we first introduce the overall OAHM generation pipeline, followed by a description of our spatially-aware coarse motion generation strategy. Finally, we detail our Motion Score Distilled Interaction (MSDI)  method, where we propose motion score distillation sampling (MSDS) and incorporate spatial and physical constraints to optimize human motion trajectories, enabling the synthesis of physically plausible and natural motion sensitive to object.

\subsection{OAHM Generation Framework}

\begin{figure*}[t]
    \centering 
    \includegraphics[width=1.0\textwidth, keepaspectratio]{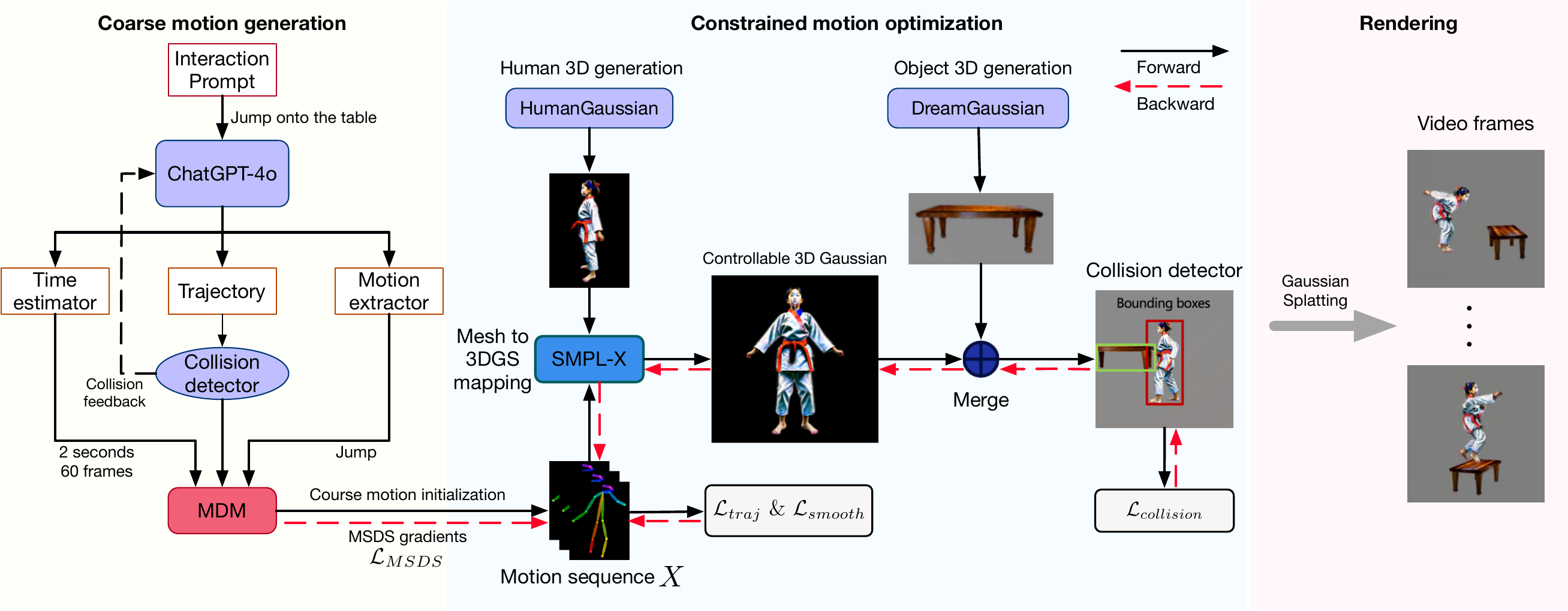}

    \caption{\textbf{Method overview.} The framework includes 4 components: human and object 3D generation, coarse trajectory generation, constrained motion optimization, and rendering. 
}
    \label{fig:method_overview_stacked}
\end{figure*}

With expressive 3D representations as the foundation, we employ HumanGaussian~\citep{Liu2024} to generate high-fidelity 3D human Gaussians $G_h$ from textual prompts, and utilize DreamGaussian~\citep{Tang2023} to synthesize 3D objects from an initial shape-e geometry~\citep{jun2023shape}, as illustrated in Figure~\ref{fig:method_overview_stacked}. Given a motion sequence $X$, we establish a correspondence between the motion trajectory and the human Gaussian points $G_h$, enabling dynamic Gaussian-based human motion.

Concretely, we initialize the SMPL-X model in a rest pose, consistent with the canonical configuration of the human Gaussian points. Each Gaussian point is mapped to the nearest barycentric coordinate on the corresponding SMPL-X mesh face. By preserving these fixed barycentric correspondences, any transformation applied to the SMPL-X mesh is faithfully propagated to the associated Gaussian points, ensuring coherent deformation of the 3D human representation.\footnote{Recalculating or interpolating linear blend skinning (LBS) weights for the Gaussian points is a viable alternative; however, we focus on the barycentric mapping approach for clarity.} Note that this mapping is differentiable, enabling gradients back propagation.

With a controllable Gaussian human and an independently generated Gaussian object co-located in the same coordinate system, we render interactive sequences via Gaussian splatting. While high-quality 3D human and object representations can be readily obtained, the availability of joint 4D human-object distributions remains limited~\citep{bhatnagar2022behave}, making it infeasible to train generative models directly on such data. To this end, we propose a zero-shot OAHM generation framework. Constructing this framework and achieving realistic object-aware human motion remains highly challenging due to the need for temporally consistent, physically plausible, and semantically appropriate interactions. Therefore, this framework presents two major challenges: (1) extracting meaningful and realistic human motion distributions, and (2) enforcing human-object interaction constraints on generated motions.

\subsection{Spatial-Aware Coarse Motion Generation}

\textbf{Motion diffusion models.} To address the first challenge, we avoid relying on video diffusion models, as prior work~\citep{bahmani20244d} has shown that distilling human motion from such models often leads to unrealistic results. Instead, we leverage dedicated human motion diffusion models (MDMs)~\citep{Tevet2022a, Shafir2023}, which are currently state-of-the-art for generating plausible human motions.

Our motion representation consists of an $N$-length motion sequence $X = \{x^i\}_{i=1}^N$, where each $x^i \in \mathbb{R}^{3 + 6 + J \times 6}$ encodes the pose parameters $\theta^i \in \mathbb{R}^{J \times 6}$, global translation $r^i \in \mathbb{R}^3$, and 6D global orientation $\gamma^i \in \mathbb{R}^6$. Other parameters, such as body shape, are omitted for simplicity. During the MDM process, the $N$-frame motion sequence $X$ is subject to $T$ steps of Gaussian noise:
\begin{equation}
    q\left(X_t \mid X_{t-1}\right) = \mathcal{N}\left(\sqrt{\alpha_t} X_{t-1}, (1-\alpha_t) I\right),
\end{equation}
where $t \in \{1, \ldots, T\}$ denotes the diffusion step and $X_T \sim \mathcal{N}(0, I)$. The MDM is trained to predict the clean motion $\hat{X}_0$ given a noisy motion $X_t$ and a text condition $c$ encoded by a CLIP-based text encoder.

\textbf{LLM-based trajectory generation.} Despite the advantages of MDMs, directly sampling from these models does not guarantee meaningful object-aware motions, as they lack explicit spatial awareness necessary for modeling relationships between humans and objects. Attempts to use guidance from 2D image and video diffusion models also failed to yield reliable spatial supervision signals. 

To overcome this, we harness the spatial reasoning capabilities of LLMs. Given the initial coordinates of the human and object, along with a textual motion instruction, the LLM generates a coarse global trajectory for the human. This LLM-derived trajectory, denoted as $r_{\mathrm{LLM}}^i$ for $i = 1, \ldots, N$, can be further refined using trajectory interpolation and collision detection, enabling the system to produce physically plausible paths, such as automatic detours around obstacles. For example, when instructed to “walk four meters toward a table two meters away,” the LLM can synthesize a motion that navigates around the object.

The LLM-generated trajectory is used to initialize the global translation in the MDM framework~\citep{Shafir2023}. With estimated time/frames and extracted pure motion prompt as two additional inputs, the MDM can yield a coarse motion sequence that incorporates spatial awareness. While the resulting motions may lack fine-grained realism, they provide a strong starting point for subsequent optimization. Detailed examples of LLM prompts are provided in the Appendix.

\subsection{Constrained Motion Optimization}

The core challenge in generating realistic object-aware human motion lies not only in producing plausible human motion, but also in enforcing physical constraints such as collision avoidance and trajectory fidelity. These challenges are not easily addressed by existing generative models. As mentioned above, two major issues must be overcome: (1) generating meaningful, in-distribution human motion sequences, and (2) ensuring these motions respect spatial and physical constraints posed by objects in the environment.

While our LLM-guided approach and MDM address the extraction of plausible motion trajectories, these solutions alone cannot guarantee realistic interactions. Specifically, directly applying the LLM-generated coarse trajectories often results in infeasible or unnatural motions, as these trajectories may violate object penetration constraints or fall outside the motion distribution captured by the pre-trained MDM. Moreover, existing diffusion models are not inherently designed to encode or enforce collision and spatial constraints.

To overcome these limitations, we introduce a constrained motion optimization framework, namely, motion score distilled interaction (MSDI), that jointly refines human motion by leveraging the prior knowledge encoded in motion diffusion models, while explicitly enforcing trajectory, smoothness, and collision-avoidance constraints.

\textbf{Motion Diffusion Score Distillation Sampling (MSDS).}  
Instead of generating human motions directly from the diffusion model, we extract the score (gradient) information from the MDM to guide the optimization of both trajectories and poses under physical constraints. Specifically, we propose \textit{Motion Diffusion Score Distillation Sampling} (MSDS), which optimizes human motion $X$ by maximizing the log-likelihood under the MDM prior. The gradient of the MSDS objective is given by:
\begin{equation}
    \nabla_X \mathcal{L}_{\mathrm{MSDS}}(\phi) \triangleq \mathbb{E}_{t, \epsilon}\left[w(t)\left(X - \operatorname{MDM}_\phi(X_t, t, c)\right)\right],
\end{equation}
where $\operatorname{MDM}_\phi$ denotes the pre-trained MDM and $w(t)$ is a weighting function over diffusion steps. This process aligns the optimized motion with the learned distribution of human poses and trajectories.

\textbf{Constrained Optimization Objectives.}  
To ensure the resulting motions are physically plausible and interact naturally with the object, we further introduce explicit constraints:
\begin{itemize}[leftmargin=5.5mm]
    \item \textbf{Trajectory Alignment.} We regularize the optimized trajectory to remain close to the LLM-generated coarse trajectory. The trajectory loss is defined as:
    \begin{eqnarray}
        \mathcal{L}_{\mathrm{traj}} & = & \lambda_{\mathrm{middle}} \cdot \sum_{i=2}^{N-1} \| r^i - r_{LLM}^i \|_2^2 + \nonumber \\
        & & \lambda_{\mathrm{end}} \cdot \sum_{i \in \{1, N\}} \| r^i - r_{LLM}^i \|_2^2,
    \end{eqnarray}
    where $\lambda_{\mathrm{middle}}$ and $\lambda_{\mathrm{end}}$ control the fidelity at middle and endpoint frames, respectively.

    \item \textbf{Motion Smoothness.} To prevent unnatural or abrupt changes in motion, we introduce a jerk (third derivative) regularization:
    \begin{equation}
        \mathcal{L}_{\mathrm{smooth}} = \sum_{i=1}^N \left\| \frac{d^3 r^i}{dt^3} \right\|_2^2,
    \end{equation}
    where, in practice, the derivative is approximated using finite differences over adjacent frames.

    \item \textbf{Collision Avoidance.} To prevent human-object penetration, we employ a two-stage collision detection and penalty scheme. First, we compute the intersection $\mathcal{C}$ of the 3D bounding boxes for the human and object. If $\mathcal{C}$ is non-empty, we evaluate pairwise collisions between object points $o_i \in \mathcal{C}$ and their nearest human points $h_j \in \mathcal{C}$. The collision loss is then given by:
    \begin{equation}
        \mathcal{L}_{\mathrm{collision}} = \max\left(\mathbf{n}_j \cdot (h_j - o_j), -\epsilon_c\right),
    \end{equation}
    where $\mathbf{n}_j$ is the normal vector at $h_j$ and $\epsilon_c$ is a collision margin hyperparameter.
\end{itemize}

\textbf{MSDI Objective.}  
The final loss function for human motion optimization combines the above terms:
\begin{eqnarray}
    \mathcal{L} & = & \lambda_{\mathrm{MSDS}} \cdot \mathcal{L}_{\mathrm{MSDS}} + \lambda_{\mathrm{traj}} \cdot \mathcal{L}_{\mathrm{traj}} + \nonumber\\ 
    & & \lambda_{\mathrm{smooth}} \cdot \mathcal{L}_{\mathrm{smooth}} 
    + \lambda_{\mathrm{collision}} \cdot \mathcal{L}_{\mathrm{collision}},
\end{eqnarray}
where $\lambda_{\mathrm{MSDS}}$, $\lambda_{\mathrm{traj}}$, $\lambda_{\mathrm{smooth}}$, and $\lambda_{\mathrm{collision}}$ are hyperparameters balancing the different objectives.

This MSDI constrained optimization aggregates all gradients to the motion sequence $X$ and updates it. Through this optimization, we ensure that the generated motions $X$ are not only realistic according to the learned motion diffusion prior but also spatially and physically consistent with the surrounding environment and objects.

\section{Experiments}
\label{sec:qualitative_results}
In this section, we evaluate the effectiveness of our proposed MSDI framework for object aware 4D human motion generation. Our experiments include both qualitative and quantitative analyses, benchmarking against the state-of-the-art 4Dfy method. We report results across a suite of objective metrics designed to assess motion realism, diversity, and physical plausibility. Ablation studies further demonstrate the importance of main components within our pipeline.

\begin{figure*}[t]
    \centering 
    \includegraphics[width=1.0\textwidth, keepaspectratio]{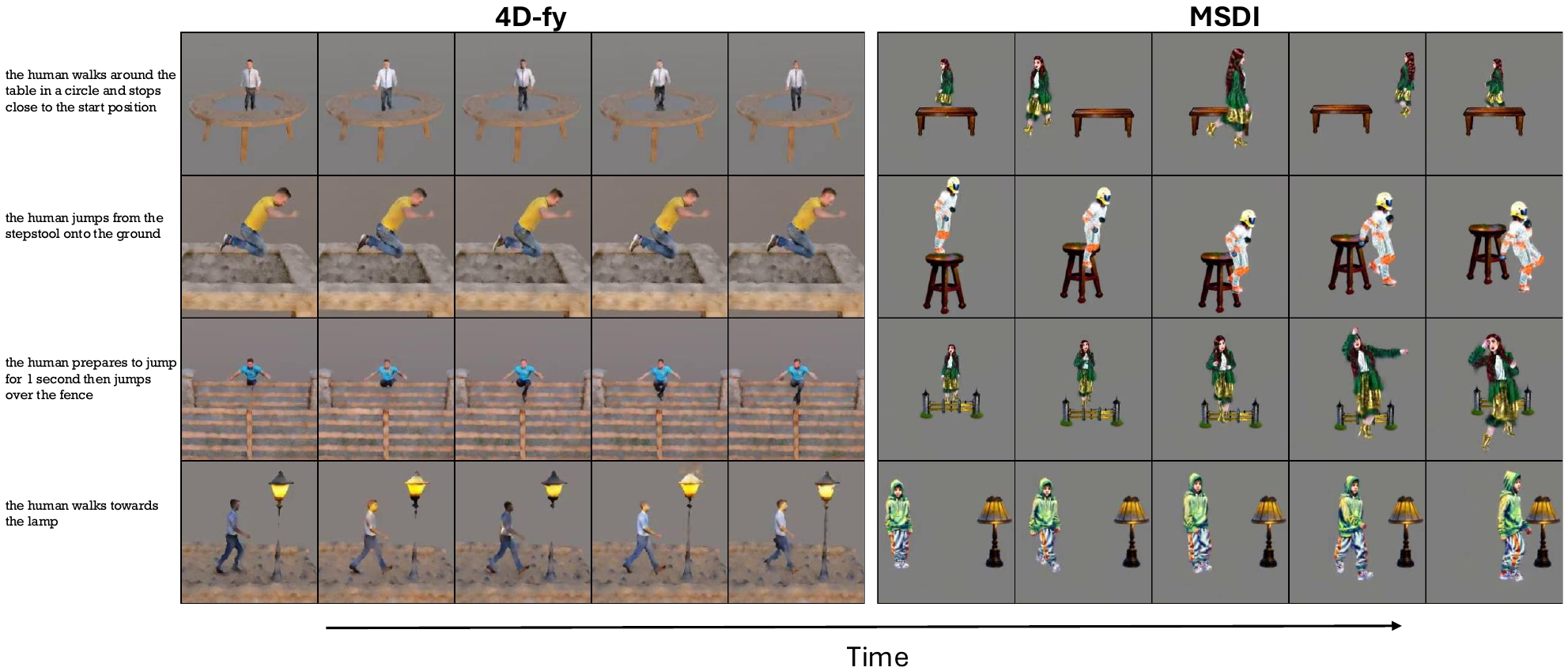}

    \caption{\textbf{Qualitative Results.}
 Generated videos from 4Dfy and MSDI across various text prompts. Each row corresponds to a different prompt. Within each row, columns display frames sampled at incremental timesteps from the generated video, illustrating temporal progression and motion characteristics. The frames are center cropped for better visibility.
}
    \label{fig:qualitative_results}
\end{figure*}

\begin{comment}
\begin{sidewaysfigure}
    \centering
    % The width here is relative to the new "page height" after rotation.
    % So, \textwidth in a sidewaysfigure often refers to the shorter dimension of the physical page.
    % \textheight often refers to the longer dimension.
    \includegraphics[width=0.9\textheight]{sec/figures/qualitative_results.pdf}
    \caption{\textbf{Qualitative Results.}
 Generated videos from 4Dfy and MSDI across various text prompts. Each row corresponds to a different prompt. Within each row, columns display frames sampled at incremental timesteps from the generated video, illustrating temporal progression and motion characteristics.
}
    \label{fig:qualitative_results}
\end{sidewaysfigure}
\end{comment}

\subsection{Metrics}
\label{sec:metrics}

To quantitatively assess the quality of generated object-aware human motion, we adapt several metrics that collectively measure pose realism, motion diversity, and temporal dynamics. We evaluate using established metrics like Optical Flow Score~\citep{liu2024evalcrafter} and we introduce three metrics designed to asses motion dynamics: Pose Plausibility, Pose Variation and Trajectory Length. We propose this suite of metrics because there is no single universally accepted metric to quantify the perceptual quality of human motion. It is important to consider these metrics in combination, as any individual metric can be trivially satisfied by a degenerate solution (e.g., a high trajectory score with a static, implausible pose). Since the metrics operate on different scales and cannot be combined arithmetically, their value lies in the holistic, comparative assessment of different methods. A model can only be judged to produce high-quality motion if it demonstrates strong and balanced performance across this entire suite.

% A significant challenge in 4D human motion generation is the lack of standardized metrics that specifically capture the quality and dynamism of spatial-temporal human motion.

For each generated video, we extract per-frame 3D human meshes using HMR2.0~\citep{goel2023humans}, which estimates SMPL parameters~\citep{SMPLX} for every detected human instance using a ViTDet detector~\citep{li2022exploring}. From these per-frame SMPL models and multi-view RGB frames, we compute the following metrics:\footnote{Further details on metrics and user study are provided in the supplementary material.}

\textbf{Pose Plausibility.}
We evaluate the realism of each human pose using VPoser~\citep{pavlakos2019expressive}, a variational autoencoder trained on large-scale pose data. For each frame $t$, we convert the predicted SMPL body pose parameters $\bm{\theta}_t^{\text{pose}}$ into VPoser-compatible axis-angle representation $\bm{\phi}_t \in \mathbb{R}^{N_V \times 3}$, then encode these to obtain a posterior $q(z_t|\bm{\phi}_t)$ over the latent pose space. The plausibility for each frame is quantified by the KL divergence to a standard normal prior $p(z)$.

%%%%%%
% Put the following into the appendix
%%%%%%
% We evaluate the realism of each human pose using VPoser~\citep{pavlakos2019expressive}, a variational autoencoder trained on a large dataset of human poses, which learns a latent representation $z$ for human body poses.
% For each frame $t$ in a generated video, we extract the SMPL body pose parameters $\bm{\theta}_t^{\text{pose}} \in \mathbb{R}^{23 \times 3 \times 3}$. These are converted to an axis-angle representation for the $N_V=21$ joints compatible with VPoser, denoted $\bm{\phi}_t \in \mathbb{R}^{N_V \times 3}$.
% We then use the VPoser encoder to obtain the parameters of a posterior distribution $q(z_t | \bm{\phi}_t)$ over the latent pose space.
% The pose plausibility score for the frame is computed as the Kullback-Leibler (KL) divergence between this encoded distribution and a standard normal prior $p(z) = \mathcal{N}(\mathbf{0}, \mathbf{I})$:
% \begin{equation}
%     \mathcal{L}_{\text{plaus}, t} = D_{KL}(q(z_t | \bm{\phi}_t) || p(z_t)).
% \end{equation}

\textbf{Pose Variation.}
To quantify diversity and motion magnitude, we measure the temporal standard deviation of the pose vector $\bm{\phi}_t$ (flattened dimension $K=N_V \times 3$) across all frames. High variation reflects diverse and dynamic motions.
%%%%%%
% Put the following into the appendix
%%%%%%
% To quantify the amount of motion and the diversity of poses throughout a sequence, we measure the temporal variation of the SMPL body pose parameters.
% Using the axis-angle pose vectors $\bm{\phi}_t \in \mathbb{R}^{K}$ (where $K = N_V \times 3 = 63$ is the flattened dimension of the VPoser-compatible pose parameters) for each frame $t=1, \dots, T$, we compute the standard deviation $\sigma_j$ for each of the $K$ pose parameters across time.

\textbf{Trajectory Length.}
To assess the extent of global character movement within the 3D space, we calculate the trajectory length of the root joint.
For each frame $t$, HMR2.0 provides the 3D keypoint coordinates. The total trajectory length is the sum of Euclidean distances between the root joint positions in consecutive frames.
A longer trajectory length suggests more substantial displacement of the character over time.

\begin{comment}
\textbf{Trajectory Length.}
To assess the extent of global character movement within the 2D image plane, we calculate the trajectory length of the root joint.
For each frame $t$, HMR2.0 provides the 2D keypoint coordinates. We extract the root joint's 2D position $\mathbf{k}_t = (x_t, y_t)$.
The total trajectory length is the sum of Euclidean distances between the root joint positions in consecutive frames:
\begin{equation}
    M_{\text{Trajectory}} = \sum_{t=1}^{T-1} ||\mathbf{k}_{t+1} - \mathbf{k}_t||_2.
\end{equation}
A longer trajectory length suggests more substantial displacement of the character over time.
\end{comment}

\textbf{Optical Flow Score.}
To quantify the amount of motion and temporal dynamics, we compute an Optical Flow Score~\citep{liu2024evalcrafter}. For each of the $N_{cv}=4$ views, we estimate dense optical flow between consecutive frames using RAFT~\citep{teed2020raft}. The score for each view is the average magnitude of these flow vectors across all pixels and frames. The final Optical Flow Score is the average of these per-view scores. A higher score signifies a more pronounced motion.

\textbf{User Study.}
To complement our quantitative metrics, we also conducted a user study to qualitatively assess the performance of our method against 4D-fy. The study was designed to measure human perception of motion quality, physical plausibility, and overall realism. We followed the human evaluation setup established by 4D-fy and MAV3D \citep{Singer2023}.

\subsection{Results}

\begin{figure*}[t]
    \centering 
    \includegraphics[width=1.0\textwidth, keepaspectratio]{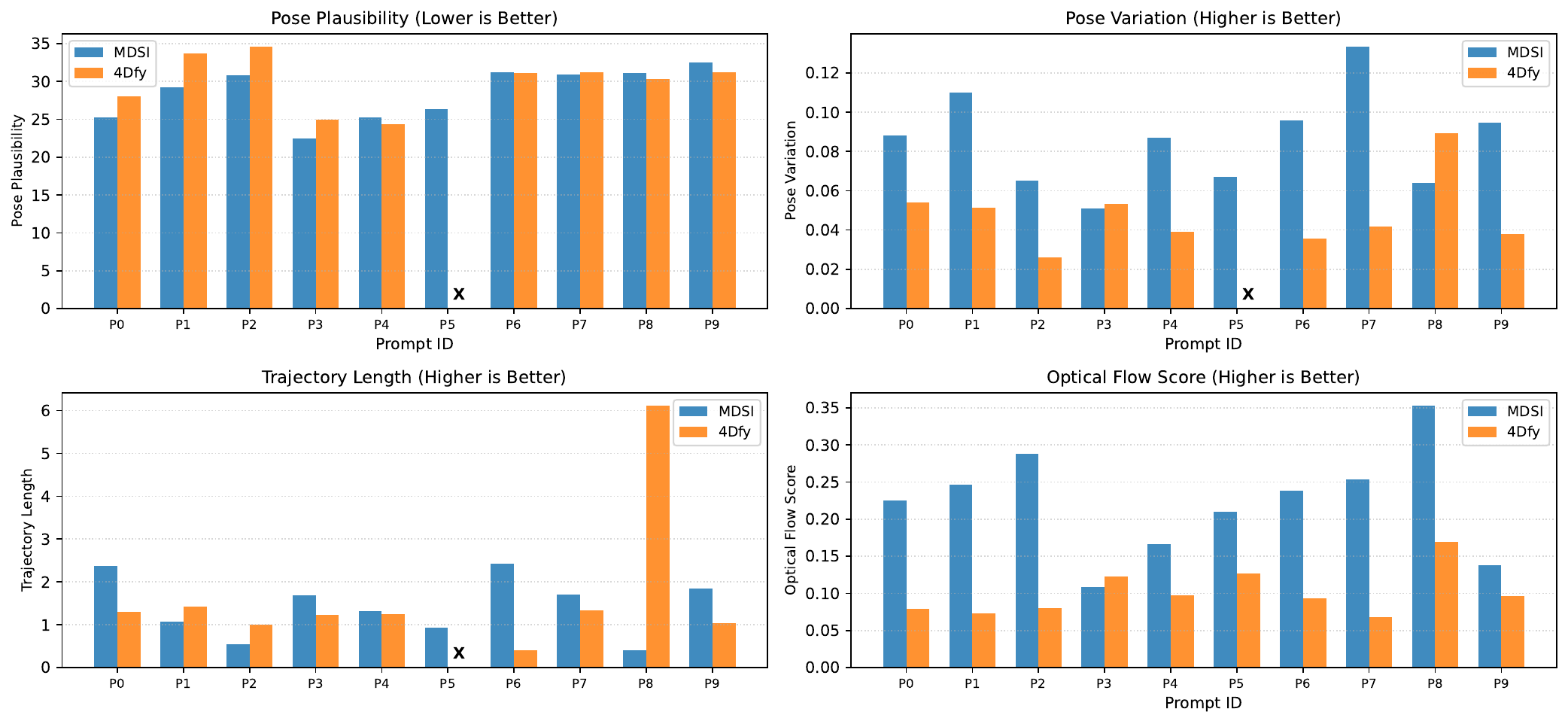}

    \caption{\textbf{Quantitative Results. } Quantitative comparison of MSDI and 4Dfy. The bar chart displays scores for 4 key metrics across 10 text prompts. An 'X' marker indicates that the metric failed to detect any humans in all four generated views for that particular prompt.}
    \label{fig:quantitative_results}
\end{figure*}

\begin{comment}
We quantitatively and qualitatively compare MSDI against 4Dfy. Quantitative scores for four metrics across ten prompts are visualized in Figure~\ref{fig:quantitative_results}

MSDI demonstrates consistently stronger performance in generating high-quality human motion. As seen in Figure~\ref{fig:quantitative_results}, our method generally achieves better \textbf{Pose Plausibility} and significantly higher \textbf{Pose Variation} and \textbf{Optical Flow Scores}. This quantitative advantage translates to qualitatively more diverse, consistent, and physically grounded human motions as seen in Figure ~\ref{fig:qualitative_results}. Our generated videos have dynamic actions and coherent movements. In contrast, 4Dfy often produces videos where frames appear largely similar, with only minor arm or leg movements and the human subject frequently remains static. It's limited human motion could also be attributed to the inherent constraints of the underlying video diffusion model, such as VideoCrafters~\citep{chen2023videocrafter1} used for Score Distillation Sampling in the current experiments. Furthermore, 4Dfy's generations sometimes suffer from visual artifacts, such as human and object attributes flickering into noise. MSDI also tends to produce more extensive spatial movements, as indicated by its frequent lead in \textbf{Trajectory Length}.

In summary, MSDI excels in producing diverse, physically plausible, and temporally consistent motions, while also demonstrating greater robustness.
\end{comment}

We conduct a comprehensive set of experiments to evaluate our method. The results demonstrate that MSDI consistently outperforms baseline 4D-fy method overall across all metrics.

\paragraph{Quantitative and Qualitative Analysis.}
As shown in our quantitative analysis (Figure~\ref{fig:quantitative_results}), MSDI shows a clear advantage. Specifically, it achieves substantially higher scores in both Pose Variation and Optical Flow, indicating diverse and larger motion. Furthermore, MSDI produces better Pose Plausibility and longer Trajectory Lengths for most prompts, while remaining comparable on others. 
% These results confirm our method's ability to generate temporally rich and plausible motion sequences.

This numerical advantage translates directly to visually perceptible improvements, as shown in our qualitative comparisons in Figure~\ref{fig:qualitative_results}. In contrast, 4Dfy often produces videos where frames appear largely similar, with only minor arm or leg movements, and the human subject frequently remains static in the same position. It's limited human motion could be attributed to the inherent constraints of the underlying video diffusion model, such as VideoCrafters~\citep{chen2023videocrafter1} used for Score Distillation Sampling. MSDI  using Motion SDS generates coherent and physically-grounded interactions with human subject showing larger change in their positions over time.

\paragraph{User Study.}
To validate that our quantitative and qualitative findings align with human judgment, we performed a formal user study comparing MSDI against 4D-fy. The results, summarized in Table~\ref{tab:user_study}, show a decisive preference for MSDI across all categories. Crucially, the preference in \textbf{Motion Quality (MQ)} is an overwhelming \textbf{87\%}, directly validating our core technical contribution. This result also confirms that the higher Pose Plausibility and Variation captured by our metrics correspond to motions that humans perceive as significantly more natural and realistic. The high preference for \textbf{Appearance (AQ) (79\%)}, \textbf{3D Structure (SQ) (71\%)} and \textbf{Text Alignment (TA) (75\%)} further suggests that physically plausible motion enhances overall visual fidelity and text alignment. The \textbf{80\%} overall preference underscores that generating believable 4D videos hinges not just on appearance, but critically on the quality of the motion itself.

\begin{table}[h!]
\centering
\setlength{\tabcolsep}{4.5pt}
\begin{tabular}{@{}lccccc@{}}
\toprule
\textbf{Preferred Method} & \textbf{AQ} & \textbf{SQ} & \textbf{MQ} & \textbf{TA} & \textbf{Overall} \\
\midrule
\textbf{MSDI} (\%) & \textbf{79\%} & \textbf{71\%} & \textbf{87\%} & \textbf{75\%} & \textbf{80\%} \\
4D-fy (\%) & 21\%  & 29\% & 13\% & 25\% & 20\% \\
\bottomrule
\end{tabular}
\caption{User study results comparing MSDI with 4D-fy. We report the percentage of times users preferred our method. MSDI significantly outperforms 4D-fy across all metrics, with all results being statistically significant ($p < 0.001$).}
\label{tab:user_study}
\end{table}

\subsection{Ablation study}

\begin{figure}[t]
    \centering 
    \includegraphics[width=0.47\textwidth, keepaspectratio]{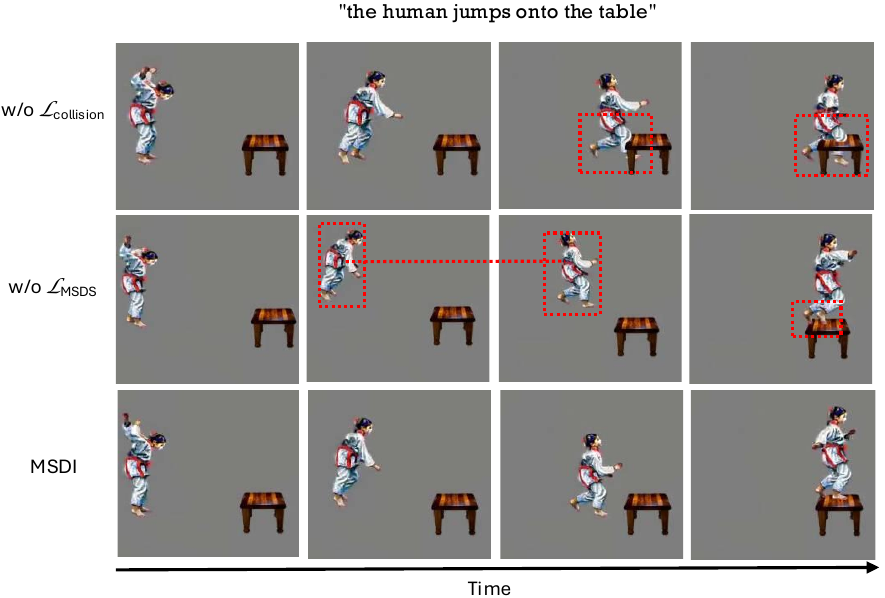}

    \caption{\textbf{Ablation study on key components of MSDI.} 
  We visualize the impact of removing our main loss terms for the prompt ''the human jumps onto the table''.}
  
    \label{fig:ablation_study}
\end{figure}

We conduct qualitative ablation studies to demonstrate the importance of key components in our proposed method. We focus on the prompt ``the human jumps onto the table'' to highlight specific failure modes when certain losses are excluded. For all variants, the high-level LLM planned trajectory remains consistent. Visualizations are provided in Figure~\ref{fig:ablation_study}.

\textbf{Effect of Collision Loss ($\mathcal{L}_{\text{collision}}$).}
Without collision loss $\mathcal{L}_{\text{collision}}$, the optimization fails to enforce physical non-penetration constraints. The generated human visibly penetrates or pierces into the table surface during the landing phase of the jump. 

\textbf{Effect of Motion Diffusion Score Distillation Sampling ($\mathcal{L}_{\text{MSDS}}$).}
Excluding the MSDS $\mathcal{L}_{\text{MSDS}}$, significantly degrades the quality of the human motion and object interaction, particularly contact points. Without $\mathcal{L}_{\text{msds}}$, the human appears to float during the jump and makes unnatural contact, with improperly planted feet.

\section{Conclusion}
\label{sec:conclusion}

In this work, we introduced Motion Score Distilled Interaction (MSDI), a novel zero-shot framework for object-aware human motion generation. Our approach uniquely combines the strengths of 3D Gaussian representations for high-fidelity visuals, motion diffusion models for realistic human movement priors, and large language models for spatial reasoning and initial trajectory planning. A key component of our framework is Motion Diffusion Score Distillation Sampling (MSDS), which allows us to refine human motion by leveraging gradients from pre-trained motion diffusion models. This, coupled with our constrained optimization strategy that considers trajectory alignment, motion smoothness, and collision avoidance, enables the generation of interactions that are not only natural but also physically plausible and respectful of object presence.

Unlike previous methods that often require extensive training on specific datasets, MSDI operates in a zero-shot manner. This means it can generalize to novel interactions without retraining, making it a scalable and adaptable solution. Our experiments have shown that MSDI can produce realistic human motions interacting with static 3D objects, overcoming common issues like unnatural distortions and physical violations seen in outputs from methods relying solely on video diffusion models. We believe MSDI offers a promising direction for creating more dynamic and believable 4D content by effectively integrating 3D physical and semantic priors into the generation process.
%%%%%%%%%%%%%%%%%%%%%%%%%%%%%%%%%%%%%%%%%%%%%%%%%%%%%%%%%%%% 
{
    \small
    \bibliographystyle{ieeenat_fullname}
    \bibliography{main}
}
\newpage
\section{Technical Appendices and Supplementary Material}
% Technical appendices with additional results, figures, graphs and proofs may be submitted with the paper submission before the full submission deadline (see above), or as a separate PDF in the ZIP file below before the supplementary material deadline. There is no page limit for the technical appendices.

\subsection{Evaluation Metrics}
\textbf{Pose Plausibility.}
A lower KL divergence indicates that the pose is more similar to those seen during VPoser's training, and thus more plausible. The final Pose Plausibility score for a video is the average $\mathcal{L}_{\text{plaus}, t}$ over all $T$ frames:
\begin{equation}
    M_{\text{Plausibility}} = \frac{1}{T} \sum_{t=1}^{T} \mathcal{L}_{\text{plaus}, t}.
\end{equation}
It is worth noting that pose plausibility utilizes a pretrained variational autoencoder, whose performance can be constrained by its original training data, potentially limiting generalization to out-of-distribution poses.

\textbf{Pose Variation.} we first compute the standard deviation $\sigma_j$ for each of the $K$ pose parameters across time:
\begin{equation}
    \sigma_j = \text{std}(\{\phi_{1,j}, \phi_{2,j}, \dots, \phi_{T,j}\}), \quad j=1, \dots, K.
\end{equation}
A higher value indicates more significant changes in pose throughout the video, suggesting more dynamic motion. The Pose Variation metric is then the mean of these standard deviations:
\begin{equation}
    M_{\text{Variation}} = \frac{1}{K} \sum_{j=1}^{K} \sigma_j.
\end{equation}

\textbf{Trajectory Length.}
To assess the extent of global character movement within the 3D space, we calculate the trajectory length of the root joint.
For each frame $t$, HMR2.0 provides the 3D keypoint coordinates. We extract the root joint's 3D position $\mathbf{k}_t = (x_t, y_t, z_t)$.
The total trajectory length is the sum of Euclidean distances between the root joint positions in consecutive frames:
\begin{equation}
    M_{\text{Trajectory}} = \sum_{t=1}^{T-1} ||\mathbf{k}_{t+1} - \mathbf{k}_t||_2.
\end{equation}
A longer trajectory length suggests more substantial displacement of the character over time.

\begin{figure}[t]
    \centering 
    \includegraphics[width=1.0\linewidth, keepaspectratio]{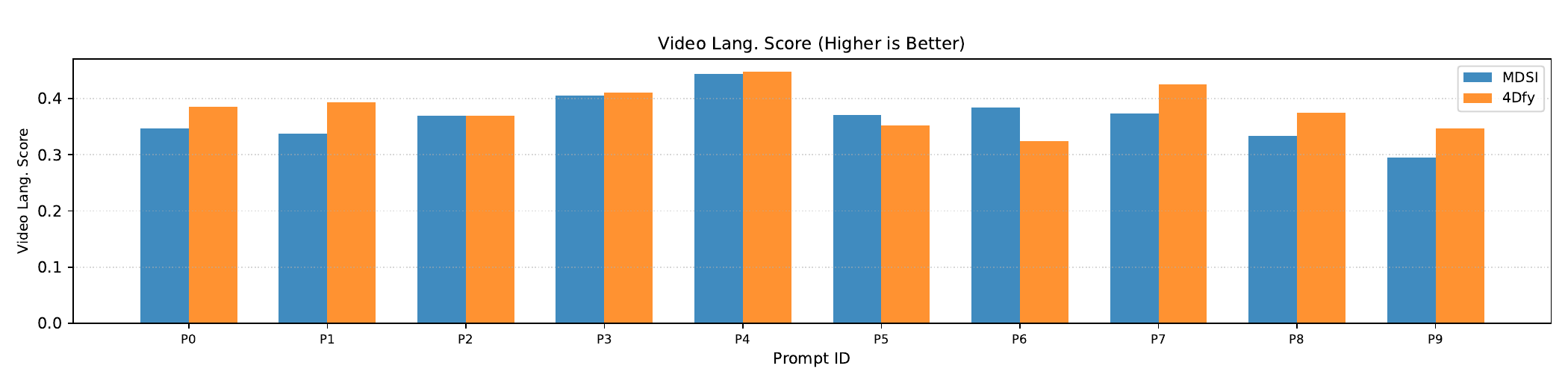}

    \caption{\textbf{Video Language Score} Comparison of MSDI and 4Dfy.}
    \label{fig:quantitative_results_video_language_scores}
\end{figure}

\textbf{Video-Language Score.}
To measure the semantic alignment between the input text prompt and the generated video, we use InternVideo2~\citep{wang2024internvideo2}, a video-text foundation model. For each of the $N_{cv}=4$ generated views, we compute the cosine similarity between the text embedding of the prompt and the video embedding. The final Video-Language Score is the average of these similarity scores across all views. A higher score indicates better prompt-video alignment.

4Dfy often achieves a higher \textbf{Video Language Score} (See Figure~\ref{fig:quantitative_results_video_language_scores}), this may stem from a bias in the metric towards static or motion-limited scenes. Consequently, the metric might prioritize overall scene-text alignment over nuanced motion quality, potentially favoring 4Dfy despite its weaker human motion dynamics. This observation is pertinent, as previous works have often relied on image-based metrics (e.g., CLIP~\citep{Radford2021} scores) for video-text alignment, which are arguably even less sensitive to temporal dynamics. Moreover, the human evaluation study shows that video generated by our method has high preference (75\%) over 4Dfy for Text Alignment (TA). This shows that the video text alignment scores using video language models does not truly capture the human perception of motion quality.

\subsection{User Study Methodology.}
We followed human evaluation methodology established by 4D-fy~\citep{bahmani20244d} and MAV3D~\citep{Singer2023}. We collected responses from 11 human evaluators. For a diverse set of 10 text prompts, each evaluator was shown a pair of videos generated by MSDI and 4D-fy. Participants were asked to choose the superior video based on five criteria:
\begin{itemize}
    \item \textbf{Appearance Quality (AQ):} The visual clarity and appeal of the generated human and object.
    \item \textbf{3D Structure Quality (SQ):} The realism and consistency of the 3D shapes across multiple viewpoints.
    \item \textbf{Motion Quality (MQ):} The naturalness, dynamism, and physical plausibility of the human's movements.
    \item \textbf{Text Alignment (TA):} How accurately the video's content reflects the input text prompt.
    \item \textbf{Overall Preference (OP):} The evaluator's subjective choice for the better video, considering all the above aspects.
\end{itemize}

\subsection{Evaluation Prompts}
\label{app:prompts}

\begin{table}
\centering
\small
\begin{tabular}{c p{0.6\linewidth}}
\toprule
\textbf{Prompt ID} & \textbf{Prompt Text} \\
\midrule
0 & the human walks around the table in a circle and stops close to the start position \\
1 & the human prepares to jump for 1 second then jumps over the fence \\
2 & the human jumps from the stepstool onto the ground \\
3 & the human walks on the clouds \\
4 & the human walks towards the lamp \\
5 & the human falls down from the stepstool \\
6 & the human crawls under the table \\
7 & the human prepares to jump for 1 second then jumps onto the table and stops on the surface of the table for 1 second \\
8 & the human falls down on the ground \\
9 & the human sits down on ground with legs cross \\
\bottomrule
\end{tabular}
\caption{List of text prompts used for evaluation.}
\label{tab:evaluation_prompts}
\end{table}

Table~\ref{tab:evaluation_prompts} lists the text prompts used for the quantitative and qualitative evaluation.

\begin{figure*}[t]
  \centering % Center the content of the figure environment

  % First figure
  \includegraphics[width=1.0\textwidth]{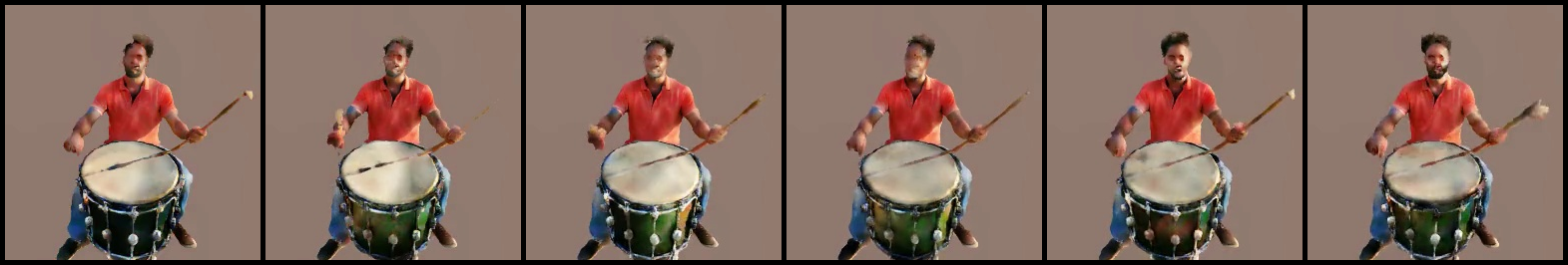} % Adjust width as needed
  \vspace{2mm}

  % Second figure
  \includegraphics[width=1.0\textwidth]{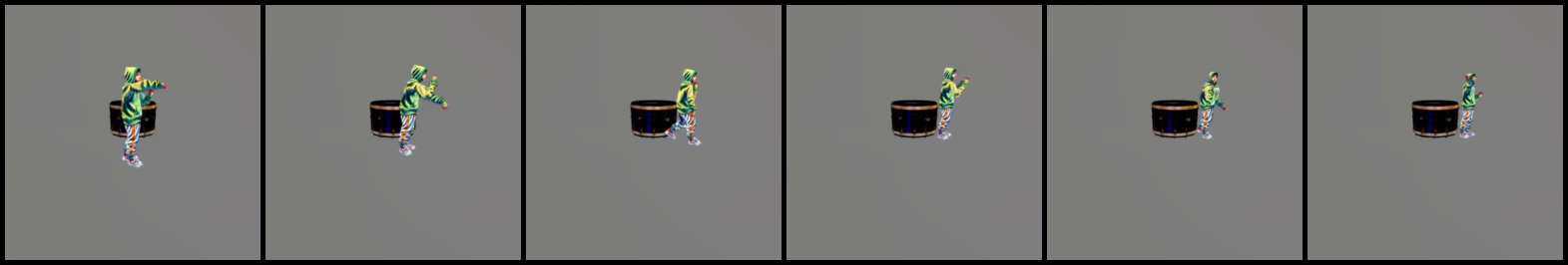} % Adjust width as needed
  
  \caption{Generated motion for the prompt: \textit{"the human is playing a drum"}. Top: 4Dfy. Bottom: MSDI}
  \label{fig:limitation_playing_drum}
\end{figure*}

\subsection{Limitations}\label{sec:limitations}

Despite its advancements, MSDI has several limitations offering avenues for future work.

First, the final output quality is tied to the pre-generated 3D assets and their initial placement and orientation. Suboptimal inputs or challenging initial setups (e.g., incorrect facing, distant objects) can hinder the generation of plausible interactions, as our framework doesn't currently optimize this initial scene layout.

Second, our reliance on LLMs for initial "coarse" trajectory generation can be a bottleneck. LLMs may produce suboptimal, physically impractical, or semantically incorrect paths for complex prompts or environments, providing a poor starting point for optimization.

Third, the framework struggles with fine-grained interactions, especially detailed hand and finger movements (e.g., realistically playing a drum, Figure~\ref{fig:limitation_playing_drum}). Current motion models and representations lack the specificity for such dexterous tasks, leading to generalized rather than precise contact.

Fourth, while our collision avoidance works for general movements, it may be less robust or efficient for highly complex object geometries or very intricate, close-quarters interactions.

Fifth, MSDI is currently designed for human interactions with static objects. Handling dynamic objects or multi-agent scenarios remains a future challenge.

Finally, the system's performance is dependent on the capabilities of the underlying pre-trained motion diffusion models, and the optimization process requires careful hyperparameter tuning to balance different objectives.

\subsection{Compute Resources}
All experiments were conducted on a system equipped with 1 NVIDIA A100 GPUs, 128 CPU cores, and 1TB of CPU memory.
Generating a single 4D video clip with 4Dfy (all three of its stages) required approximately 24 hours. MSDI completed the generation of human and object artifacts followed by the optimization process in approximately 5 hours per prompt using the same computational resources.

\subsection{Multi View Qualitative Results}

\begin{figure*}[htbp] % htbp are placement suggestions: here, top, bottom, page
  \centering % Center the content of the figure environment

  % First figure
  \includegraphics[width=1.0\textwidth]{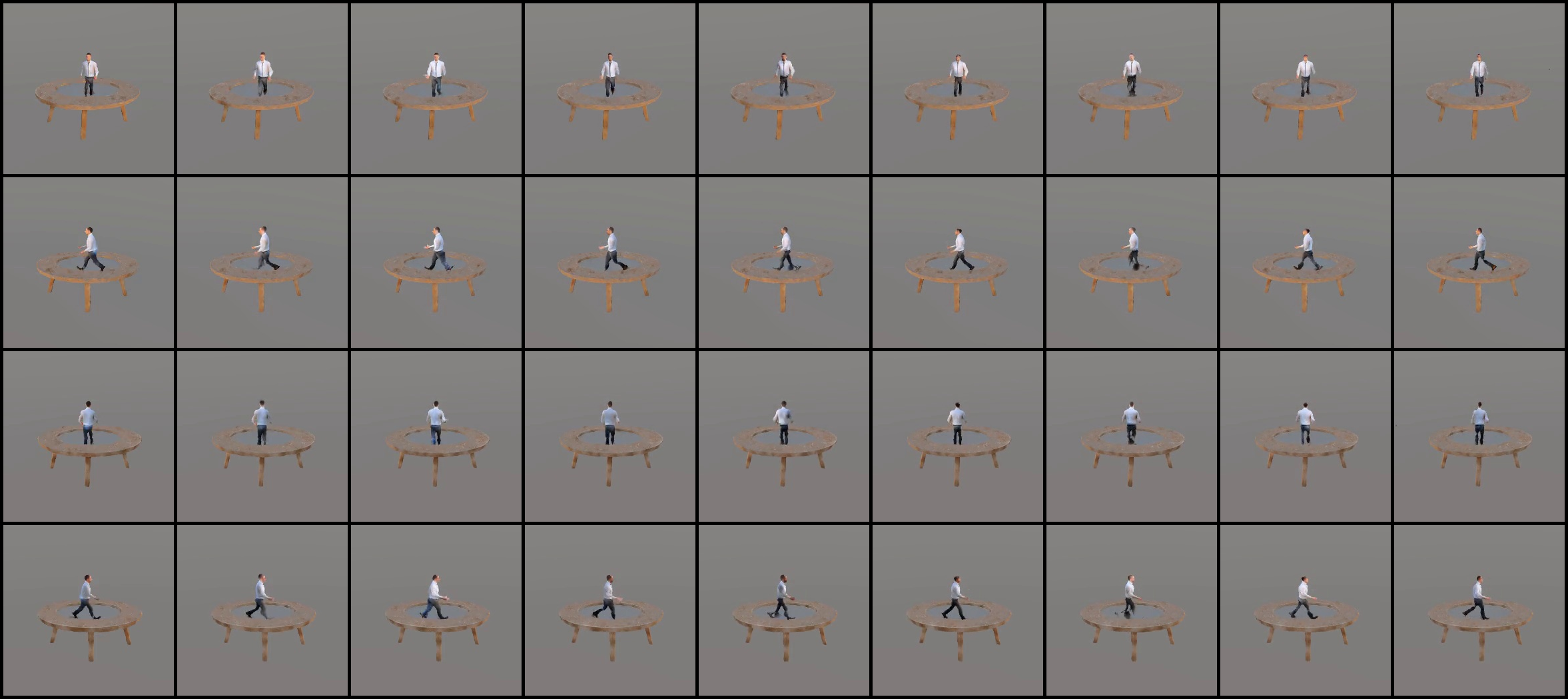} % Adjust width as needed
  \vspace{2mm}

  % Second figure
  \includegraphics[width=1.0\textwidth]{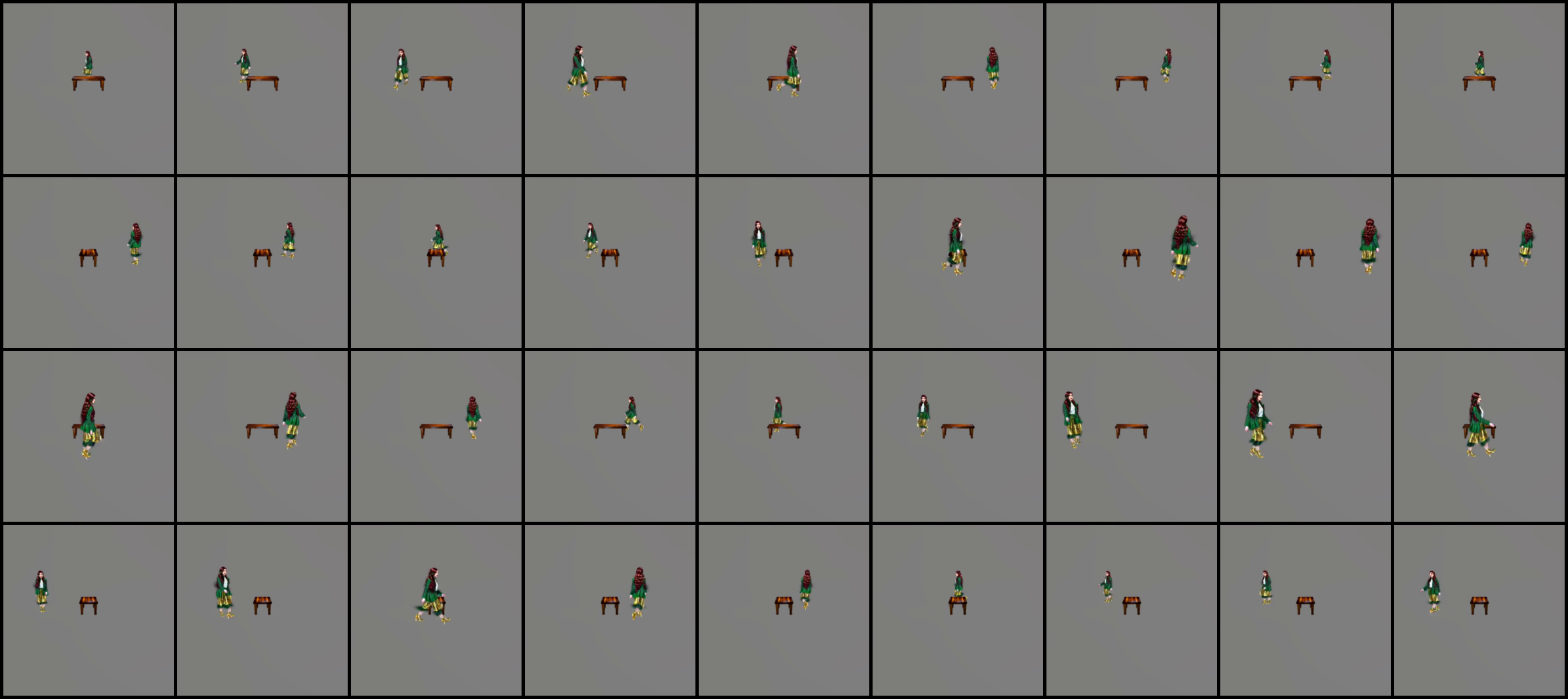} % Adjust width as needed
  
  \caption{Generated motion for the prompt: \textit{"a human walks around a table in a circle and stops close to the start position"}. Top: 4Dfy. Bottom: MSDI. The four rows illustrate the motion from different camera viewpoints}
  \label{fig:prompt0_mulitview}
\end{figure*}

\begin{figure*}[htbp] % htbp are placement suggestions: here, top, bottom, page
  \centering % Center the content of the figure environment

  % First figure
  \includegraphics[width=1.0\textwidth]{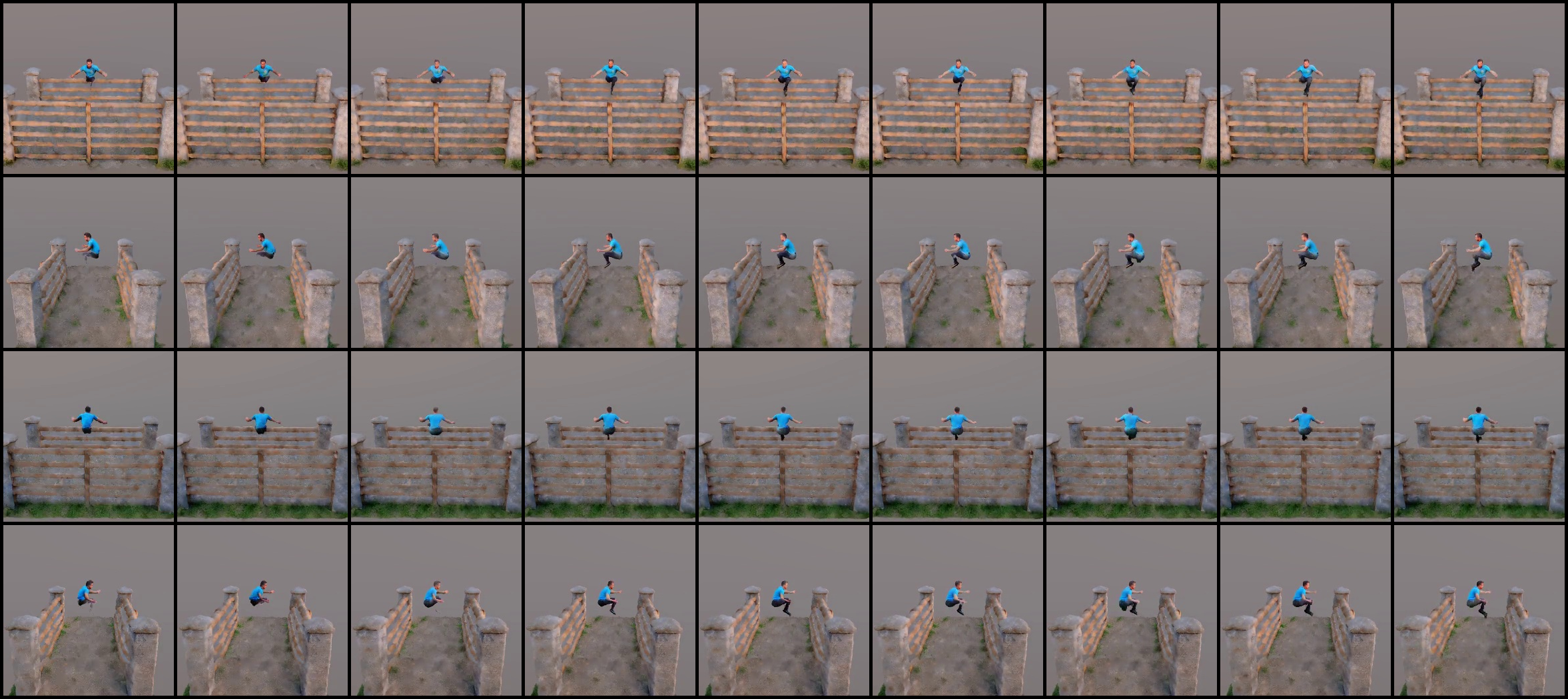} % Adjust width as needed
  \vspace{2mm}

  % Second figure
  \includegraphics[width=1.0\textwidth]{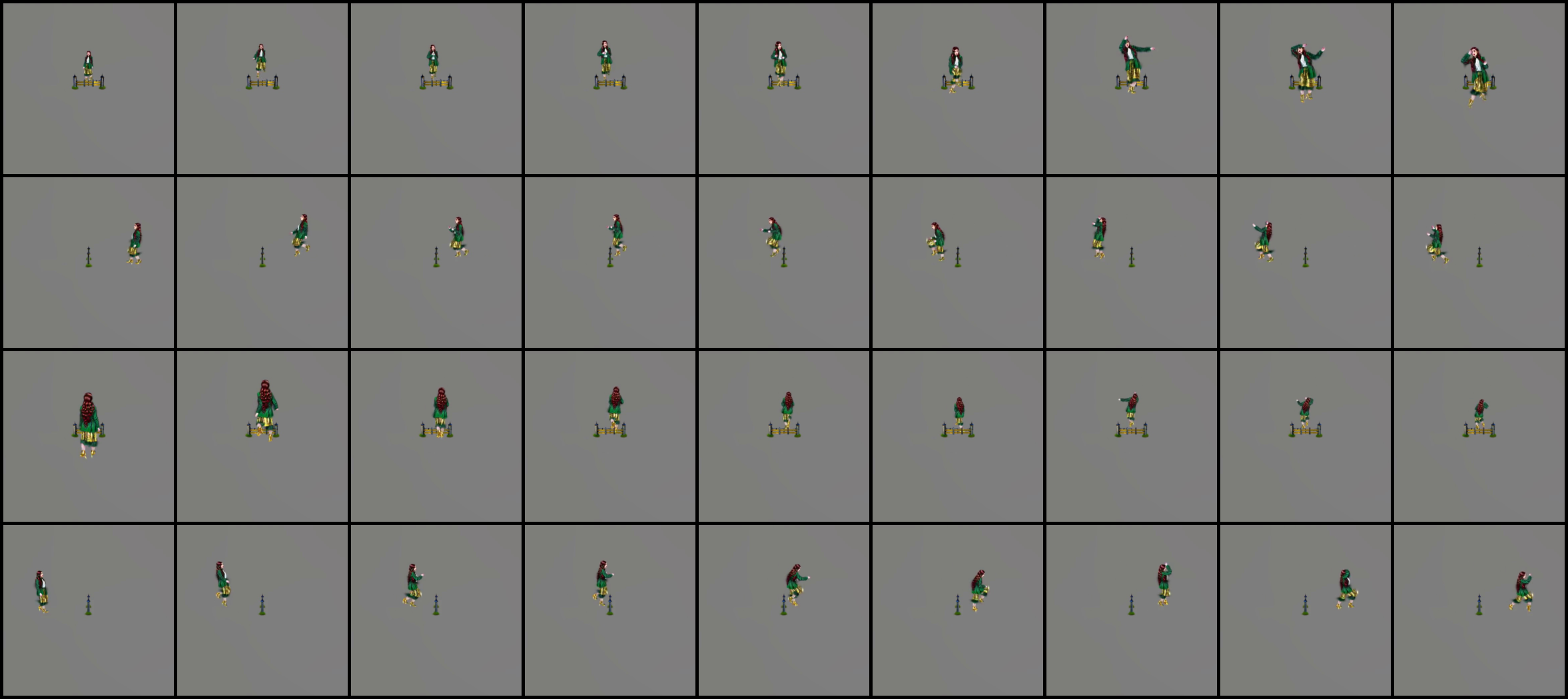} % Adjust width as needed
  
  \caption{Generated motion for the prompt: \textit{"the human prepares to jump for 1 second then jumps over the fence"}. Top: 4Dfy. Bottom: MSDI. The four rows illustrate the motion from different camera viewpoints}
  \label{fig:prompt1_mulitview}
\end{figure*}

\begin{figure*}[htbp] % htbp are placement suggestions: here, top, bottom, page
  \centering % Center the content of the figure environment

  % First figure
  \includegraphics[width=1.0\textwidth]{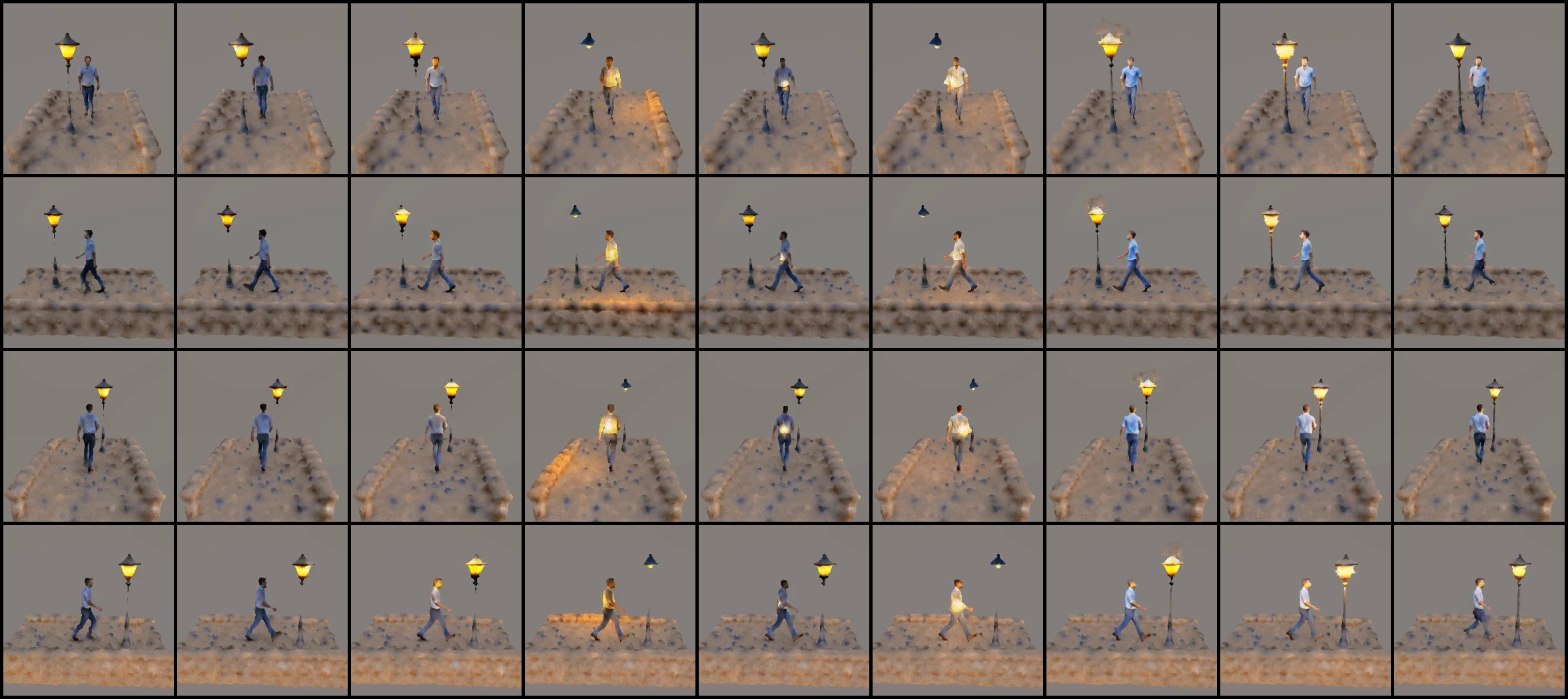} % Adjust width as needed
  \vspace{2mm}

  % Second figure
  \includegraphics[width=1.0\textwidth]{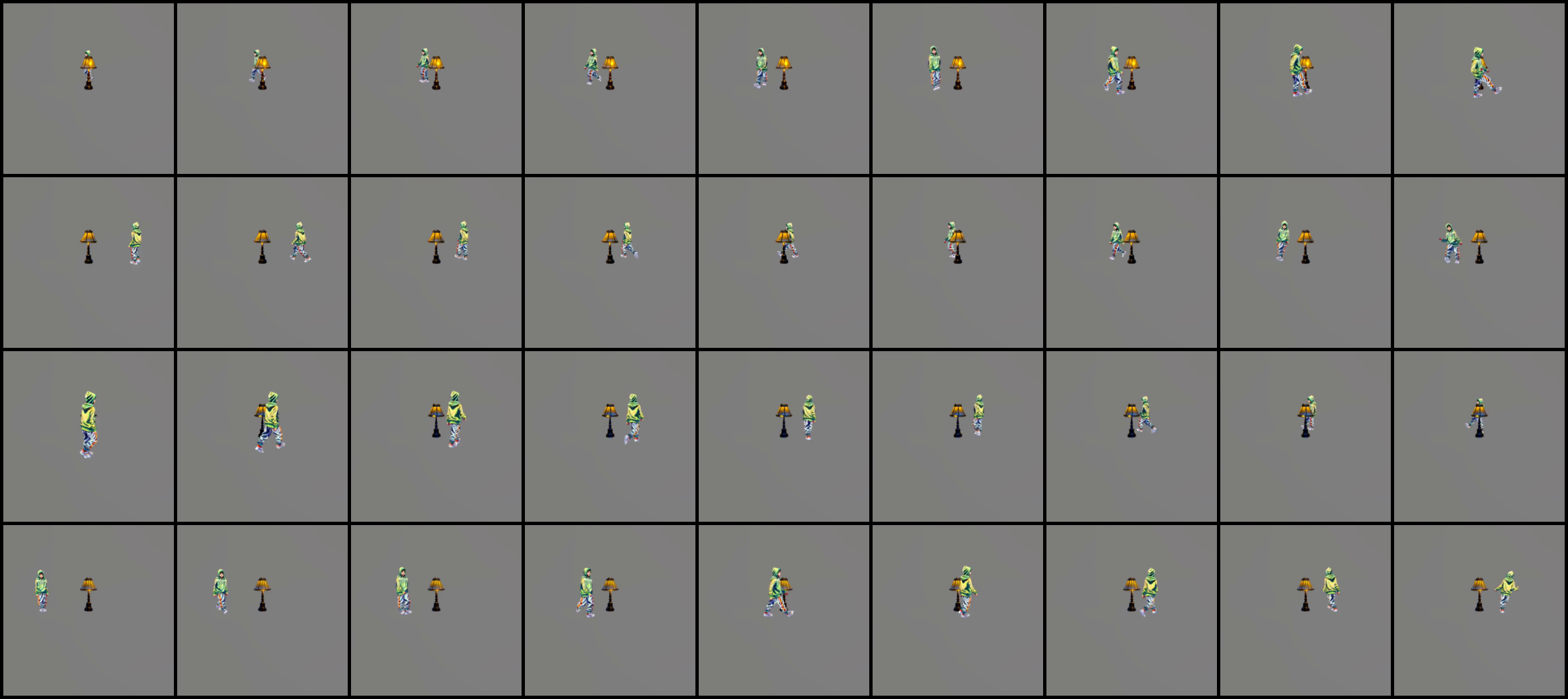} % Adjust width as needed
  
  \caption{Generated motion for the prompt: \textit{"the human walks towards the lamp"}. Top: 4Dfy. Bottom: MSDI. The four rows illustrate the motion from different camera viewpoints}
  \label{fig:prompt4_mulitview}
\end{figure*}

\begin{figure*}[htbp] % htbp are placement suggestions: here, top, bottom, page
  \centering % Center the content of the figure environment

  % First figure
  \includegraphics[width=1.0\textwidth]{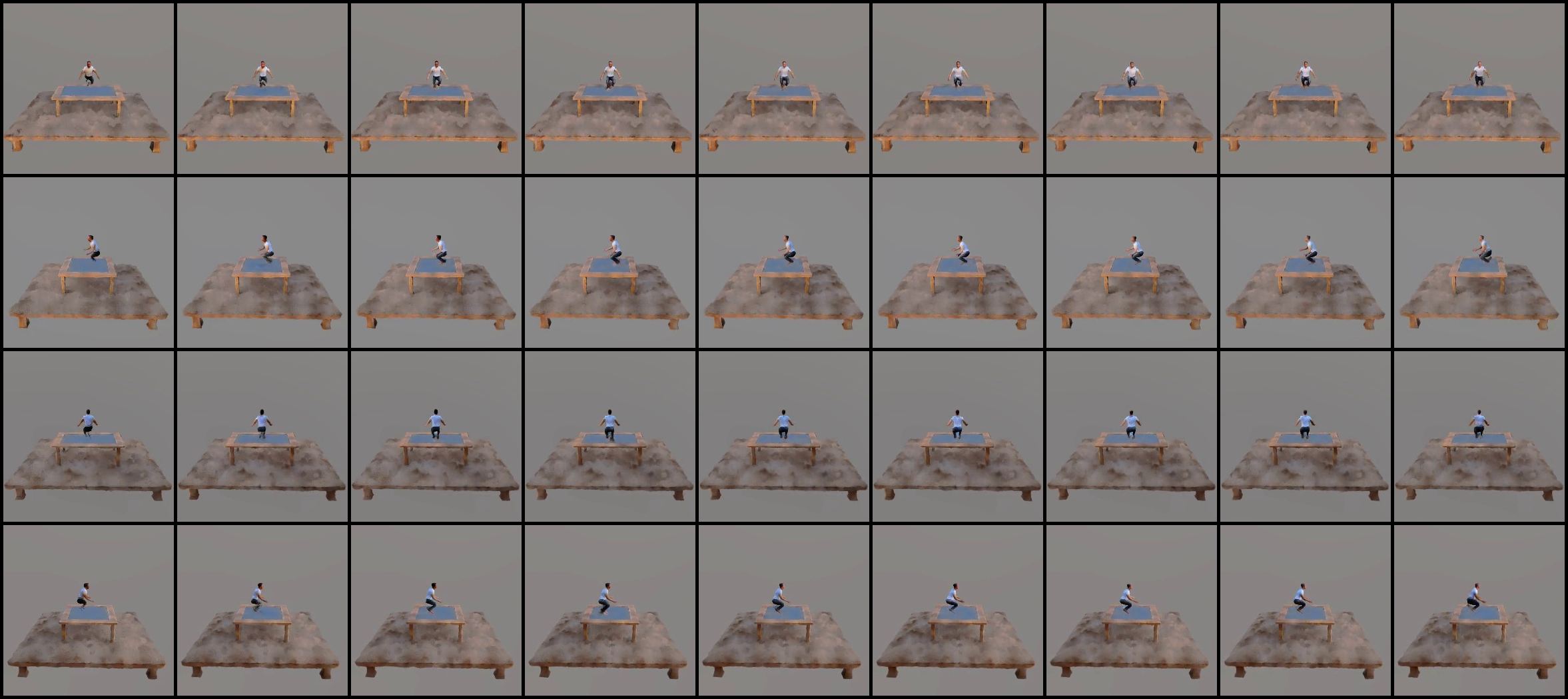} % Adjust width as needed
  \vspace{2mm}

  % Second figure
  \includegraphics[width=1.0\textwidth]{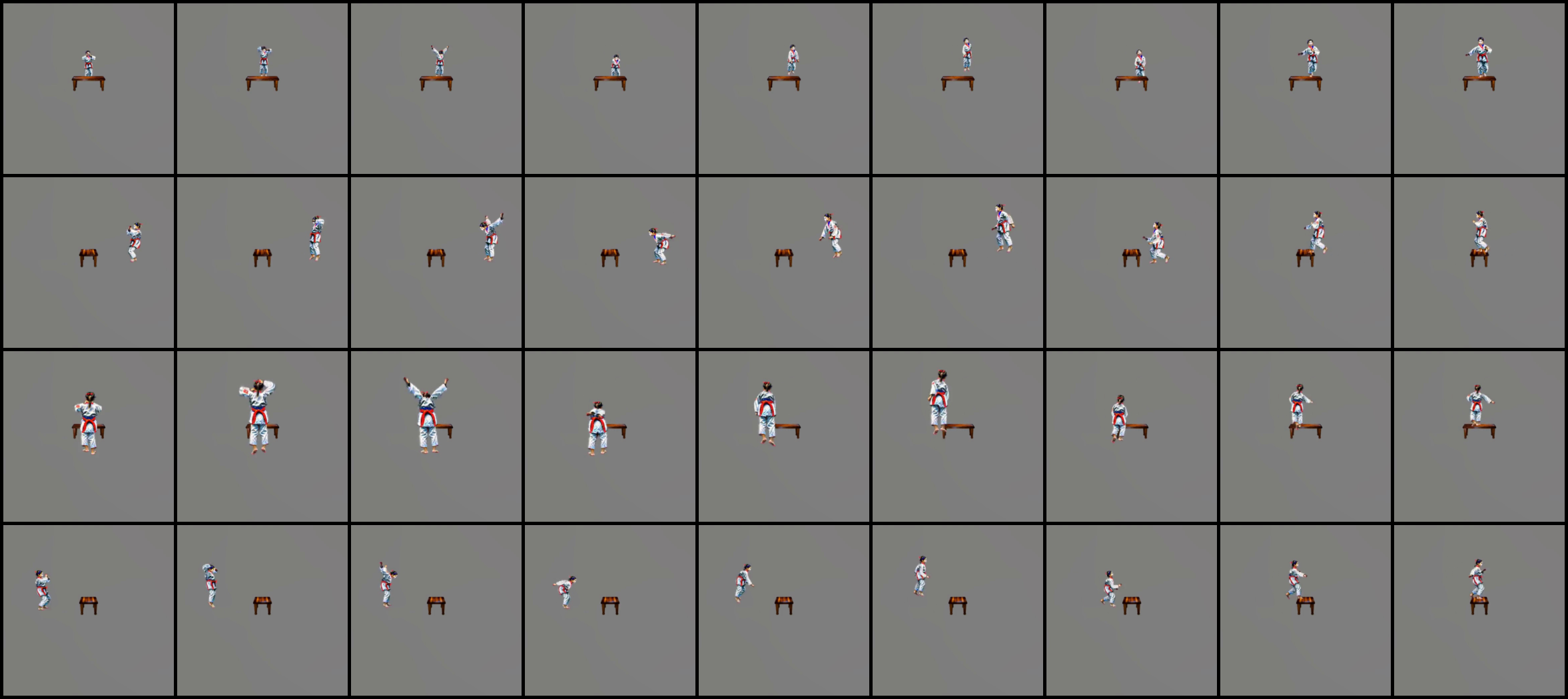} % Adjust width as needed
  
  \caption{Generated motion for the prompt: \textit{"the human prepares to jump for 1 second then jumps onto the table and stops on the surface of the table for 1 second"}. Top: 4Dfy. Bottom: MSDI. The four rows illustrate the motion from different camera viewpoints}
  \label{fig:prompt7_mulitview}
\end{figure*}

Figures ~\ref{fig:prompt0_mulitview}, ~\ref{fig:prompt1_mulitview}, ~\ref{fig:prompt4_mulitview}, ~\ref{fig:prompt7_mulitview}, shows comparison of generated motion with 4Dfy and MSDI from different camera angles.

\end{document}